\documentclass{article}

\usepackage{arxiv}

\usepackage[utf8]{inputenc} 
\usepackage[T1]{fontenc}    
\usepackage{hyperref}       
\usepackage{url}            
\usepackage{booktabs}       
\usepackage{amsfonts}       
\usepackage{nicefrac}       
\usepackage{microtype}      
\usepackage{lipsum}

\usepackage{bm}
\usepackage{amsmath}
\usepackage{amssymb}
\usepackage{amsfonts}
\usepackage{graphicx}
\def\vec#1{\mathbf{#1}}
\usepackage{algorithm}
\usepackage{algorithmic}
\usepackage{comment}

\usepackage{mathtools}
\usepackage{proof}
\def\qed{\Box}

\usepackage{amsthm}
\theoremstyle{plain}
\newtheorem{theorem}{Theorem}[section]

\newtheorem{lemma}[theorem]{Lemma}

\theoremstyle{definition}
\newtheorem{definition}[theorem]{Definition}

\theoremstyle{remark}

\title{Meta-learning of semi-supervised learning from tasks with heterogeneous attribute spaces}

\author{
  Tomoharu Iwata\\
  NTT Communication Science Laboratories\\
  \And
  Atsutoshi Kumagai\\
  NTT Computer and Data Science Laboratories\\  
}
\date{}

\begin{document}
\maketitle

\begin{abstract}
We propose a meta-learning method for semi-supervised learning that learns from multiple tasks with heterogeneous attribute spaces. The existing semi-supervised meta-learning methods assume that all tasks share the same attribute space, which prevents us from learning with a wide variety of tasks. With the proposed method, the expected test performance on tasks with a small amount of labeled data is improved with unlabeled data as well as data in various tasks, where the attribute spaces are different among tasks. The proposed method embeds labeled and unlabeled data simultaneously in a task-specific space using a neural network, and the unlabeled data's labels are estimated by adapting classification or regression models in the embedding space. For the neural network, we develop variable-feature self-attention layers, which enable us to find embeddings of data with different attribute spaces with a single neural network by considering interactions among examples, attributes, and labels. Our experiments on classification and regression datasets with heterogeneous attribute spaces demonstrate that our proposed method outperforms the existing meta-learning and semi-supervised learning methods.
\end{abstract}

\section{Introduction}

Although deep learning can achieve high predictive performance,
it requires a sufficient number of labeled data.
Semi-supervised learning and meta-learning are
machine learning approaches that improve performance with a limited number of labeled data.
In semi-supervised learning, both labeled and unlabeled data 
are used for finding class boundaries in a task~\cite{zhu2005semi}.
In meta-learning, data in various tasks are used for learning how to learn
in each task~\cite{schmidhuber:1987:srl}.
Recently, meta-learning methods for semi-supervised learning
have been proposed, where both the unlabeled data
and the data in different tasks are used~\cite{liu2018learning,ren2018meta,wang2020meta,liu2021semi,li2020online,xiao2021semi}.
However, they cannot use data from tasks in heterogeneous attribute spaces,
which denote that
their attribute spaces are different among tasks~\cite{zhu2011heterogeneous,moon2017completely,zhou2014hybrid}.
The existing methods require that the attribute spaces of
all the tasks be identical,
restricting learning from a wide variety of tasks,
which might contain useful knowledge for unseen tasks.

\begin{figure*}[t!]
  \centering
  \begin{tabular}{cc}
  \includegraphics[height=11.5em]{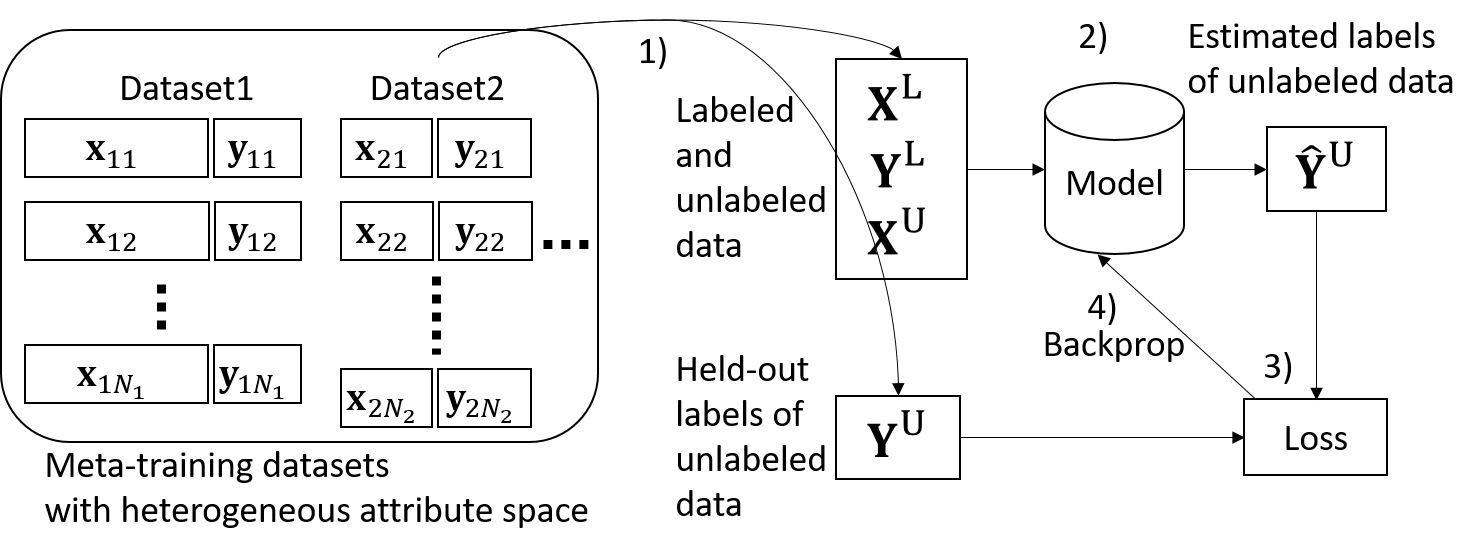} &
  \includegraphics[height=11.5em]{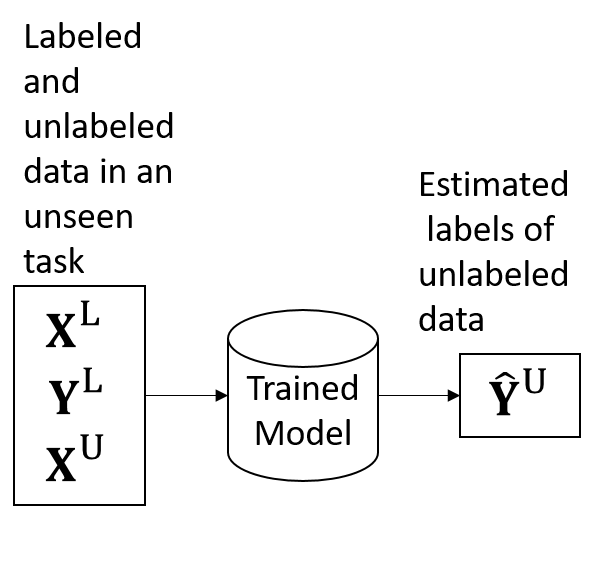}\\
  (a) Meta-training phase & (b) Meta-test phase\\
  \end{tabular}
  \caption{(a) Procedures of the proposed method in the meta-training phase. We are given meta-training datasets with heterogeneous attribute spaces. For each meta-learning step, 1) randomly sample labeled and unlabeled data $\vec{X}^{\mathrm{L}}, \vec{Y}^{\mathrm{L}}, \vec{X}^{\mathrm{U}}$ and held-out labels of unlabeled data $\vec{Y}^{\mathrm{U}}$ from a randomly selected dataset in the meta-training datasets, where the numbers of attributes and classes can be different across meta-learning steps. 2) Estimate labels of unlabeled data $\hat{\vec{Y}}^{\mathrm{U}}$ with our model, where the neural networks in our model are shared among all tasks. 3) Calculate loss between estimated $\hat{\vec{Y}}^{\mathrm{U}}$ and held-out labels $\vec{Y}^{\mathrm{U}}$ of unlabeled data. 4) Update the neural network parameters by backpropagating the loss. (b) Procedures in the meta-test phase. We are given a small number of labeled and unlabeled data in an unseen task. Using the trained model, we estimate labels of the unlabeled data.}
  \label{fig:algorithm}
\end{figure*}

This paper proposes a meta-learning method for semi-supervised learning
that can meta-learn from tasks with heterogeneous attribute spaces.
Figure~\ref{fig:algorithm} shows the meta-learning procedures of the proposed method.
In the meta-training phase, we are given meta-training datasets from various tasks, where
the attribute spaces and class spaces are different across tasks.
Our neural network-based model takes the labeled and unlabeled data in a task as input,
and outputs the estimated labels of the unlabeled data.
The neural network is shared across all tasks,
which enables us to extract common knowledge from various tasks
and use it for unseen tasks.
For each of the meta-learning steps,
our model is updated by backpropagating the loss
on the held-out labels of the unlabeled data in the meta-training datasets,
such that the generalization performance is improved.
In the meta-test phase, 
we are given a small number of labeled and unlabeled data for each unseen task,
which is a semi-supervised setting.

For handling data with heterogeneous attribute spaces in our model,
we propose variable-feature self-attention layers (VSA).
Self-attention layers have been successfully used for learning representations from sets~\cite{bahdanau2015neural,vaswani2017attention,lee2019set,kossen2021self}.
However, the existing self-attention layers require that the elements in sets
be represented by feature vectors with a fixed size.
Therefore, they cannot be used for data with heterogeneous attribute spaces.
On the other hand, VSA can perform attention even when
the feature vector sizes are varied across sets.
In our model, labeled and unlabeled data are embedded in a task-specific space
by iterating attention across examples and
attention across attributes and labels based on VSA.
Then, labels are estimated 
by adapting prototypical classification or Gaussian process regression models
in the embedding space.

The following are the main contributions of this paper:
1) To the best of our knowledge, our work is the first meta-learning method for semi-supervised learning from tasks with heterogeneous attribute spaces.
2) We propose attention layers that can handle sets with different feature vector sizes across sets.
3) We experimentally confirm that the proposed method outperforms the existing meta-learning and semi-supervised learning methods.


\section{Related work}

Many meta-learning methods have been proposed~\cite{schmidhuber:1987:srl,bengio1991learning,finn2017model,vinyals2016matching,snell2017prototypical,garnelo2018conditional,Liu_2021_ICCV}.
However, most assume that all tasks share the same attribute space.
Although heterogeneous meta-learning methods~\cite{iwata2020meta,brinkmeyer2019chameleon}
can handle heterogeneous spaces,
they are not semi-supervised methods and cannot use unlabeled data.
The proposed method is an embedding-based meta-learning
method~\cite{vinyals2016matching,snell2017prototypical,ren2018meta,bertinetto2018meta,iwata2021few},
where neural networks shared among tasks are used to embed examples,
and classification or regression models are adapted in the embedding space.
Unlike existing embedding-based methods,
the proposed method finds a task-specific embedding space
using both of the labeled and unlabeled examples in the task.
Transductive propagation networks~\cite{liu2018learning} and
semi-supervised prototypical networks~\cite{ren2018meta} are embedding-based meta-learning methods
for semi-supervised learning,
where unlabeled data are used for adaptation in the embedding space
with label propagation and soft clustering.
They cannot handle heterogeneous attribute spaces.
Heterogeneous attribute spaces have been considered in transfer learning~\cite{moon2017completely,zhu2011heterogeneous,wang2009heterogeneous,yan2017learning,li2013learning,wang2011heterogeneous,zhou2014hybrid},
which transfers knowledge in a source task to a target task.
However, these transfer learning methods require a target task for training.
On the other hand, our proposed method does not use target tasks for training neural networks.

Attention layers improved meta-learning performance~\cite{kim2018attentive,xu2020metafun}
since they can model interactions between examples.
However, these existing attention-based meta-learning methods
cannot handle heterogeneous spaces.
The existing self-attention layers~\cite{vaswani2017attention}
can be incorporated into the heterogeneous meta-learning method~\cite{iwata2020meta}.
In our experiments, we demonstrate that the proposed VSA layers
achieve better performance than such a combination of the existing methods.

\section{Proposed method}

In Section~\ref{sec:problem}, we formulate our meta-learning problem for semi-supervised learning.
In Section~\ref{sec:vssa}, we propose a
variable-feature self-attention (VSA) layer that is used in our model.
In Section~\ref{sec:model}, we present our model to embed labeled and unlabeled data
with heterogeneous attribute spaces using VSA layers.
In Section~\ref{sec:classifier}, we describe the classifiers that output the estimated labels of unlabeled data
given the embeddings.
In Section~\ref{sec:meta-learning}, we explain the meta-learning procedures of our model.
We also present the proposed method in the case of regression tasks in Section~\ref{sec:regression}.

\subsection{Problem formulation}
\label{sec:problem}

In the meta-training phase, we are given $T$ labeled meta-training datasets
with heterogeneous attribute spaces
$\mathcal{D}=\{\{(\vec{x}_{tn},\vec{y}_{tn})\}_{n=1}^{N_{t}}\}_{t=1}^{T}$,
where $\vec{x}_{tn}\in\mathbb{R}^{M_{t}}$ is the attribute vector of the $n$th example
in the $t$th dataset, $\vec{y}_{tn}\in\{0,1\}^{C_{t}}$ is the onehot vector that indicates its class label,
$N_{t}$ is the number of examples,
$M_{t}$ is the number of attributes, and $C_{t}$ is the number of classes.
The attributes and classes
(and their numbers)
can be different across datasets;
$M_{t}\neq M_{t'}$, $C_{t}\neq C_{t'}$.
Although we explain the proposed method assuming that
each dataset consists of labeled examples for simplicity,
unlabeled examples can also be contained in the meta-training datasets.

In the meta-test phase,
we are given a small number of labeled data
$\vec{X}^{\mathrm{L}}$, $\vec{Y}^{\mathrm{L}}$
and unlabeled data $\vec{X}^{\mathrm{U}}$ in an unseen test task
that are different from but related to the meta-training datasets.
Here, 
$\vec{X}^{\mathrm{L}}=[\vec{x}^{\mathrm{L}}_{1},\dots,\vec{x}^{\mathrm{L}}_{N^\mathrm{L}}]^{\top}\in\mathbb{R}^{N^\mathrm{L}\times M}$ is the attribute matrix of the labeled data,
$\vec{Y}^{\mathrm{L}}=[\vec{y}^{\mathrm{L}}_{1},\dots,\vec{y}^{\mathrm{L}}_{N^\mathrm{L}}]^{\top}\in\mathbb{R}^{N^\mathrm{L}\times C}$ is the label matrix of the labeled data,
$\vec{X}^{\mathrm{U}}=[\vec{x}^{\mathrm{U}}_{1},\dots,\vec{x}^{\mathrm{U}}_{N^{\mathrm{U}}}]^{\top}\in\mathbb{R}^{N^{\mathrm{U}}\times M}$ is the attribute matrix of the unlabeled data,
$N^{\mathrm{L}}$ is the number of labeled data,
$M$ is the number of attributes,
$C$ is the number of classes,
and $N^{\mathrm{U}}$ is the number of unlabeled data.
Our aim is to improve the classification performance on the given unlabeled data in the test task.

\subsection{Variable-feature self-attention layers}
\label{sec:vssa}

\begin{figure}[t!]
  \centering
  \includegraphics[width=23em]{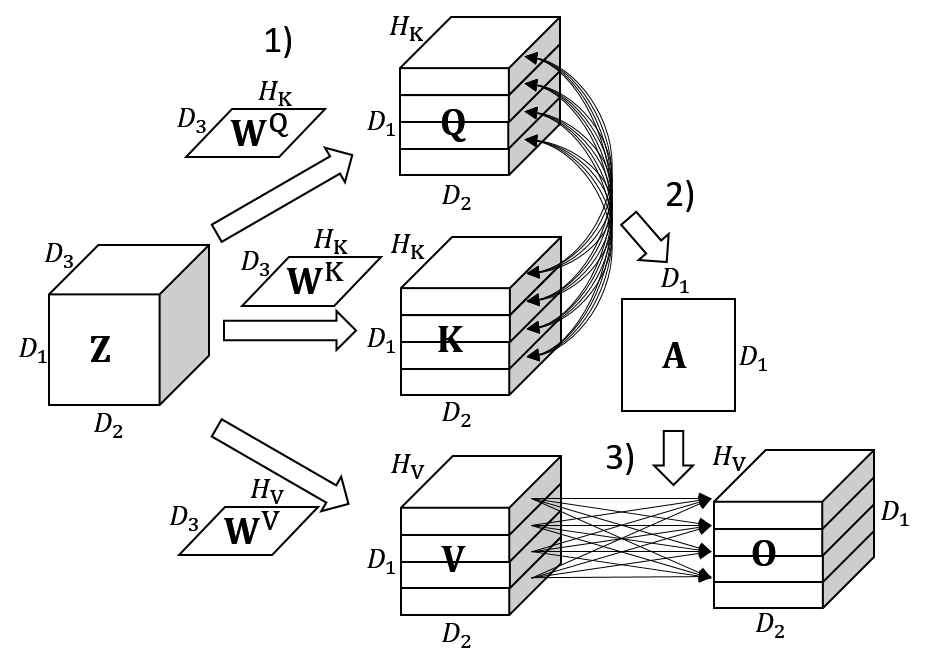}
  \caption{Variable-feature self-attention layer. 1) Input tensor $\vec{Z}$ is transformed into query $\vec{Q}$, key $\vec{K}$, and value $\vec{V}$ tensors by linear projection matrices $\vec{W}^{\rm{Q}}$, $\vec{W}^{\rm{K}}$, and $\vec{W}^{\rm{V}}$. 2) Attention weight matrix $\vec{A}$ between slices along the first mode is calculated. 3) Output tensor $\vec{O}$ is obtained
by aggregating over values $\vec{V}$ by attention weights $\vec{A}$.}
  \label{fig:attention_model}
\end{figure}


Since existing attention layers~\cite{bahdanau2015neural,vaswani2017attention,lee2019set,kossen2021self} cannot handle data with different feature sizes,
we propose variable-feature self-attention (VSA) layers
that can handle data with different feature sizes as well as
different example sizes.

Let $\vec{Z}\in\mathbb{R}^{D_{1} \times D_{2} \times D_{3}}$
be an input three-mode tensor, where
attention is performed across $D_{1}$ slices along the first mode.
The VSA layer can take a tensor as input
with different sizes of first and second modes $D_{1}$ and $D_{2}$.
For example, $D_{1}$ is the number of elements in a set,
and $D_{2}$ is the number of features.
The size of the third mode, $D_{3}$, needs to be common
for all the input tensors.
The input can be matrices by setting $D_{3}=1$.
Figure~\ref{fig:attention_model} illustrates the VSA layer.

First, input tensor $\vec{Z}$
is transformed into query $\vec{Q}\in\mathbb{R}^{D_{1}\times D_{2} \times H_{\mathrm{K}}}$,
key $\vec{K}\in\mathbb{R}^{D_{1}\times D_{2} \times H_{\mathrm{K}}}$,
and value $\vec{V}\in\mathbb{R}^{D_{1}\times D_{2} \times H_{\mathrm{V}}}$ tensors
with the mode-three product,
\begin{align}
  \vec{Q}=\vec{Z}\times_{3}\vec{W}^{\mathrm{Q}},
  \quad
  \vec{K}=\vec{Z}\times_{3}\vec{W}^{\mathrm{K}},
  \quad
  \vec{V}=\vec{Z}\times_{3}\vec{W}^{\mathrm{V}},
  \label{eq:qkv}
\end{align}
where
$\vec{W}^{\mathrm{Q}}\in\mathbb{R}^{H_{\mathrm{K}}\times D_{3}}$,
$\vec{W}^{\mathrm{K}}\in\mathbb{R}^{H_{\mathrm{K}}\times D_{3}}$,
and
$\vec{W}^{\mathrm{V}}\in\mathbb{R}^{H_{\mathrm{V}}\times D_{3}}$
are linear projection matrices,
and
$\times_{n}$ is a mode-$n$ product, e.g.,
$(\vec{Z}\times_{3}\vec{W})_{d_{1}d_{2}h}=\sum_{d_{3}=1}^{D_{3}}z_{d_{1}d_{2}d_{3}}w_{hd_{3}}$.
Since the VSA layer considers projection
from a $D_{3}$-dimensional space,
and model parameters
$\vec{W}^{\mathrm{Q}}$, $\vec{W}^{\mathrm{K}}$, and $\vec{W}^{\mathrm{V}}$
do not depend on $D_{1}$ and $D_{2}$,
the VSA layer can handle tensors with different $D_{1}$ and $D_{2}$.
On the other hand,
the existing attention layers consider projection from
a $D_{2}$-dimensional feature space
and cannot handle data with variable features.

Second,
attention weight matrix $\vec{A}\in\mathbb{R}^{D_{1}\times D_{1}}$
between slices along the first mode is calculated by
\begin{align}
  \vec{A}=\mathrm{softmax}(\vec{Q}_{(1)}\vec{K}_{(1)}^{\top}/\sqrt{D_{2}H_{\mathrm{K}}}),
  \label{eq:A}
\end{align}
where
$\vec{Q}_{(n)}$ is the mode-$n$ matricization of tensor $\vec{Q}$,
$(\vec{Q}_{(1)}\vec{K}_{(1)}^{\top})_{ij}=\sum_{d_{2}=1}^{D_{2}}\sum_{h=1}^{H_{\mathrm{K}}}q_{id_{2}h}k_{jd_{2}h}$,
and $\mathrm{softmax}$ is a softmax function normalized for each row.

Third, output tensor $\vec{O}\in\mathbb{R}^{D_{1}\times D_{2} \times H_{V}}$ is obtained
by aggregating over values $\vec{V}$ by attention weights $\vec{A}$,
\begin{align}
  \vec{O}=\vec{V}\times_{1}\vec{A}\equiv \mathrm{VSA}(\vec{Z}),
\end{align}
or 
$\vec{O}_{d_{1}d_{2}h}=\sum_{d'_{1}=1}^{D_{1}}a_{d_{1}d'_{1}}v_{d_{1}'d_{2}h}$.

The multi-head variable-feature self-attention (MVSA) layer uses
a concatenation of $R$ independent VSA layers,
\begin{align}
  \mathrm{MVSA}(\vec{Z})=\mathrm{concat}(\mathrm{VSA}_{1}(\vec{Z}),\dots,\mathrm{VSA}_{R}(\vec{Z}))\times_{3}\vec{W}^{\mathrm{O}},
  \label{eq:mvssa}
\end{align}
where
$\vec{W}^{\mathrm{O}}\in\mathbb{R}^{H\times RH_{\mathrm{K}}}$ is a linear projection matrix,
$\mathrm{concat}$ is the concatenation in the third mode,
$\mathrm{VSA}_{r}$ is the $r$th VSA layer,
and the model parameters are different across different VSAs.

VSA and MVSA layers are permutation-equivariant along the first and second modes,
i.e.,
any permutation of slices along the first (second) modes of the input tensor
permutes the slices along the first (second) modes of the output tensor.
The proof is given in Appendix~\ref{app:proof}.
The complexity of the VSA layer is
$O(D_{1}^{2}D_{2}(H_{\mathrm{K}}+H)+D_{1}D_{2}D_{3}(H_{\mathrm{K}}+H_{\mathrm{V}}))$.
When the size of the second mode of the input tensor is one $D_{2}=1$,
the VSA layer corresponds to the standard attention layer.

\subsection{Embedding models of labeled and unlabeled data with heterogeneous attribute spaces}
\label{sec:model}

\begin{figure*}[t!]
  \centering
  \includegraphics[width=43em]{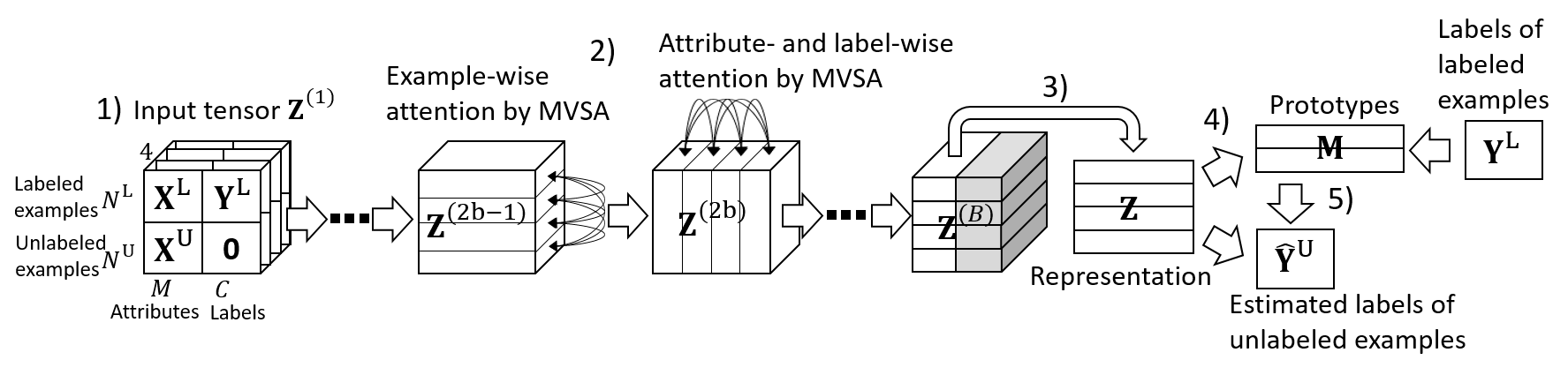}
  \caption{Our model. 1) Labeled and unlabeled data $\vec{X}^{\mathrm{L}}, \vec{Y}^{\mathrm{L}}, \vec{X}^{\mathrm{U}}$ are transformed into tensor $\vec{Z}^{(1)}$ as in Eq.~(\ref{eq:z1}). 2) Tensor $\vec{Z}^{(b)}$ is updated by iterating multi-head variable-feature self-attention along examples and along attributes and labels. 3) Embeddings for examples are obtained. 4) Prototypes for classes are calculated using the embeddings of the labeled data. 5) Labels of the unlabeled data are estimated based on the distance from the prototypes in the embedding space.}
  \label{fig:model}
\end{figure*}

Given labeled data $\vec{X}^{\mathrm{L}}$, $\vec{Y}^{\mathrm{L}}$
and unlabeled data $\vec{X}^{\mathrm{U}}$ in a task,
our neural network-based model obtains embeddings of the labeled and unlabeled examples
$\{\vec{z}^{\mathrm{L}}_{n}\}_{n=1}^{N^{\mathrm{L}}}$ and
$\{\vec{z}^{\mathrm{U}}_{n}\}_{n=1}^{N^{\mathrm{U}}}$, 
where $\vec{z}^{\mathrm{L}}_{n},\vec{z}^{\mathrm{U}}_{n}\in\mathbb{R}^{H}$.
Our model is shared across all tasks,
which can have different numbers of attributes $M$, classes $C$, and examples $N^{\mathrm{L}}, N^{\mathrm{U}}$.
By model sharing, we can extract knowledge from various tasks
on learning the embeddings of labeled and unlabeled data for semi-supervised learning.
Some existing meta-learning methods also
use neural networks with the same parameters
across different tasks for obtaining
task-specific embeddings~\cite{garnelo2018conditional,garnelo2018neural,kim2019attentive,iwata2020meta}.
However, these existing methods cannot embed
labeled and unlabeled data with heterogeneous attribute spaces.

First, we construct
input three-mode tensor $\vec{Z}^{(1)}\in\mathbb{R}^{(N^{\mathrm{L}}+N^{\mathrm{U}})\times (M+C) \times 4}$
that contains information on the labeled and unlabeled data
as shown in 1) in Figure~\ref{fig:model}.
In particular,
the slices of $\vec{Z}^{(1)}$ along the third mode are given by
\begin{align}
  \vec{Z}^{(1)}_{::1}=
  \begin{bmatrix}
    \vec{X}^{\mathrm{L}} & \vec{Y}^{\mathrm{L}}\\
    \vec{X}^{\mathrm{U}} & \vec{0}_{N^{\mathrm{U}} C}\\
  \end{bmatrix},
  \quad
  \vec{Z}^{(1)}_{::2}=
  \begin{bmatrix}
    \vec{1}_{N^{\mathrm{L}} M} & \vec{1}_{N^{\mathrm{L}} C}\\
    \vec{1}_{N^{\mathrm{U}} M} & \vec{0}_{N^{\mathrm{U}} C}\\
  \end{bmatrix},
  \quad
  \vec{Z}^{(1)}_{::3}=
  \begin{bmatrix}
    \vec{1}_{N^{\mathrm{L}} M} & \vec{0}_{N^{\mathrm{L}} C}\\
    \vec{1}_{N^{\mathrm{U}} M} & \vec{0}_{N^{\mathrm{U}} C}\\
  \end{bmatrix},
  \quad
  \vec{Z}^{(1)}_{::4}=
  \begin{bmatrix}
    \vec{0}_{N^{\mathrm{L}} M} & \vec{1}_{N^{\mathrm{L}} C}\\
    \vec{0}_{N^{\mathrm{U}} M} & \vec{1}_{N^{\mathrm{U}} C}\\
  \end{bmatrix},
  \label{eq:z1}
\end{align}
where
$\vec{0}_{N M}$ ($\vec{1}_{N M}$) represents a matrix of size $N\times M$ in which all the elements are zero (one),
and $\vec{Z}^{(1)}_{::1},\vec{Z}^{(1)}_{::2},\vec{Z}^{(1)}_{::3},\vec{Z}^{(1)}_{::4}\in\mathbb{R}^{(N^{\mathrm{L}}+N^{\mathrm{U}})\times (M+C)}$.
The first slice along the third mode
$\vec{Z}^{(1)}_{::1}$
contains information on attributes and labels, where
zero is padded for labels of the unlabeled data.
The second slice $\vec{Z}^{(1)}_{::2}$
indicates whether the element is observed or not.
The third
$\vec{Z}^{(1)}_{::3}$
and fourth slices
$\vec{Z}^{(1)}_{::4}$
indicate whether they are attributes or labels.
Concatenating indicators 
enables a neural network to transform different types of variables differently~\cite{lipton2016modeling}. 

Next, we embed the labeled and unlabeled data
in a task-specific space by alternately iterating attention across examples and
attention across attributes and labels
with MVSA layers, as shown in 2) in Figure~\ref{fig:model}.
We use the following neural network $f^{(b)}$ as a block,
\begin{align}
  f^{(b)}(\vec{Z}^{(b)})&=\vec{Z}^{(b)}\times_{3}\vec{W}^{\mathrm{R}(b)}
  +\mathrm{FF}^{(b)}(\mathrm{LN}^{(b)}(\mathrm{MVSA}^{(b)}(\vec{Z}^{(b)}))),
  \label{eq:f}
\end{align}
where
$\vec{Z}^{(b)}\in\mathbb{R}^{(N^{\mathrm{L}}+N^{\mathrm{U}})\times(M+C)\times H^{(b)}}$
is the representation at the $b$th block,
$\vec{W}^{\mathrm{R}(b)}\in\mathbb{R}^{H^{(b)}\times H^{(b+1)}}$ is a residual linear projection matrix,
$\mathrm{FF}^{(b)}:\mathbb{R}^{H^{(b)}}\rightarrow \mathbb{R}^{H^{(b+1)}}$
is a feed-forward neural network
that takes each mode-three fiber $\vec{Z}_{nm:}$ as input,
$\mathrm{LN}^{(b)}$ is a layer normalization,
and $\mathrm{MVSA}^{(b)}$ is a MVSA layer in Eq.~(\ref{eq:mvssa}).
Residual blocks and layer normalization
have been used in existing attention layers~\cite{bahdanau2015neural,vaswani2017attention,lee2019set}.
The embeddings at the even number blocks
are obtained by example-wise attention along the first mode of the tensor,
\begin{align}
  \vec{Z}^{(2b)}=f^{(2b-1)}(\vec{Z}^{(2b-1)}),
  \label{eq:z2b}
\end{align}
where examples are aligned along the first mode of $\vec{Z}^{(b)}$.
The embeddings at the odd number blocks
are obtained by
attribute- and label-wise attention along the second mode of the tensor
by swapping the first and second modes of the tensor
before and after applying $f$,
\begin{align}
  \vec{Z}^{(2b+1)}=f^{(2b)}(\vec{Z}^{(2b)\top})^{\top},
  \label{eq:z2b1}
\end{align}
where $\top$ represents the transposition of the first and second modes,
and attributes and labels are aligned along the second mode of $\vec{Z}^{(b)}$.
The approach of alternately obtaining example and attribute representations
has been used~\cite{iwata2020meta,kossen2021self}.
By iterating the attention,
we can find embeddings by considering
interactions among examples, attributes, and labels.
The interactions between labeled and unlabeled examples are important
to estimate labels of the unlabeled examples using similarities to the labeled examples.
The interactions between attributes and labels help
discover attributes that are related to labels.
The interactions between attributes help extract information of
the joint distribution of the attributes that characterizes data even without labels.

As the embedding for each example,
we use the output of the last $B$th block,
$\vec{z}_{n}^{\mathrm{L}}=[\vec{z}^{(B)}_{n1:},\dots,\vec{z}^{(B)}_{nM:}],\vec{z}_{n}^{\mathrm{U}}=[\vec{z}^{(B)}_{N^{\mathrm{L}}+n,1,:},\dots,\vec{z}^{(B)}_{N^{\mathrm{L}}+n,M,:}]\in\mathbb{R}^{H}$
where
$H=MH^{(B)}$ is the dimension of the embedding space,
$\vec{z}^{(B)}_{nm:}\in\mathbb{R}^{H^{(B)}}$ is the mode-three fiber of the $B$th output on the $m$th attribute of the
$n$th example,
and only the elements on the attributes over second mode $m=1,\cdots,M$ are used
as shown at 3) in Figure~\ref{fig:model}.

Our model is equivariant on the permutation over labeled examples,
over unlabeled examples,
over attributes, and over classes
since the MVSA layers are permutation-equivariant 
along the first and second modes.
This property is desirable since
their orders should not affect the embeddings.
Note that our model is not equivariant on the permutation between the labeled and unlabeled examples
since we use binary matrix $\vec{Z}^{(1)}_{::2}$ in Eq.~(\ref{eq:z1}) to specify whether
examples are labeled or unlabeled.
Similarly,
our model is not equivariant on the permutation between attributes and classes
due to $\vec{Z}^{(1)}_{::2}$, $\vec{Z}^{(1)}_{::3}$, and $\vec{Z}^{(1)}_{::4}$.
More explanations on our embedding models are described in Appendix~\ref{app:embedding}.

\subsection{Classification given embeddings}
\label{sec:classifier}

We adapt a classifier to the given labeled data in the embedding space.
We use prototypical classifiers~\cite{snell2017prototypical}
for predicting the class labels given representations
$\{\vec{z}_{n}^{\mathrm{L}}\}_{n=1}^{N^{\mathrm{L}}}$,
$\{\vec{z}_{n}^{\mathrm{U}}\}_{n=1}^{N^{\mathrm{U}}}$,
as shown in 4) and 5) in Figure~\ref{fig:model},
where
the class probability is estimated with a Gaussian mixture model
in the embedding space adapted to the labeled data.
We can use other classifiers that are used for meta-learning,
such as linear models~\cite{bertinetto2018meta}, Gaussian processes~\cite{snellbayesian,iwata2021few},
and label propagation~\cite{liu2018learning}.
The mean vector of class $c$ in the embedding space is calculated by
averaging the embeddings over the labeled examples with class $c$,
$\bm{\mu}_{c}=\frac{1}{N^{\mathrm{L}}_{c}}\sum_{n:y^{\mathrm{L}}_{nc}=1}\vec{z}^{\mathrm{L}}_{n}\in\mathbb{R}^{H}$,
where $N^{\mathrm{L}}_{c}=\sum_{n:y^{\mathrm{L}}_{nc}=1}1$ is the number of labeled examples with class $c$.
The probability of class $c$ of unlabeled example
$\vec{z}_{n}^{\mathrm{U}}$
is given using the distance to the class mean,
\begin{align}
  p(y|\vec{z}_{n}^{\mathrm{U}};\vec{X}^{\mathrm{L}},\vec{Y}^{\mathrm{L}},\vec{X}^{\mathrm{U}},\bm{\Theta})=\frac{\exp(-\parallel\vec{z}_{n}^{\mathrm{U}}-\bm{\mu}_{y}\parallel^{2})}{\sum_{c=1}^{C}\exp(-\parallel\vec{z}_{n}^{\mathrm{U}}-\bm{\mu}_{c}\parallel^{2})},
  \label{eq:py}
\end{align}
where $\bm{\Theta}$
is the parameters of our layers $f^{(1)},\dots,f^{(B-1)}$.
Labeled data $\vec{X}^{\mathrm{L}}, \vec{Y}^{\mathrm{L}}$ unlabeled data $\vec{X}^{\mathrm{U}}$, and
parameters $\bm{\Theta}$ are explicitly included in Eq.~(\ref{eq:py})
to indicate that the class probability depends on them. Note that
parameters $\bm{\Theta}$ are shared across tasks,
but $\vec{X}^{\mathrm{L}}, \vec{Y}^{\mathrm{L}}, \vec{X}^{\mathrm{U}}$ are task-specific.

\subsection{Meta-learning procedures}
\label{sec:meta-learning}

Parameters $\bm{\Theta}$ of our model
are optimized such that the expected test classification loss is minimized,
\begin{align}
  \arg\min_{\bm{\Theta}}\mathbb{E}\left[-\frac{1}{N^{\mathrm{U}}}
  \sum_{n=1}^{N^{\mathrm{U}}}\log p(y_{n}^{\mathrm{U}}|\vec{z}^{\mathrm{U}}_{n};\vec{X}^{\mathrm{L}},\vec{Y}^{\mathrm{L}},\vec{X}^{\mathrm{U}},\bm{\Theta})\right],  
  \label{eq:thetahat}
\end{align}
where
$\mathbb{E}$ represents the expectation over tasks with
attributes $\vec{X}^{\mathrm{L}}$ and labels $\vec{Y}^{\mathrm{L}}$ of the labeled data
and attributes $\vec{X}^{\mathrm{U}}$ and held-out labels $\vec{Y}^{\mathrm{U}}$ of the unlabeled data,
and $y_{n}^{\mathrm{U}}$ is the held-out class label of the $n$th unlabeled data.
By Eq.~(\ref{eq:thetahat}), we can obtain model parameters that can embed labeled and unlabeled data such that the test performance is improved when classified in the task-specific embedding space.

Algorithm~\ref{alg:train} shows the meta-learning procedures of our model.
The expectation in Eq.~(\ref{eq:thetahat}) is approximated by the Monte Carlo method
by randomly sampling datasets, labeled and unlabeled examples
in Lines 3--5.
For simplicity, we explained
the proposed method assuming that all the examples in the meta-training datasets are labeled.
The proposed method can also use unlabeled examples in meta-training datasets
by skipping the unlabeled examples from the calculation of the test loss.

The time complexity for each meta-learning step linearly increases with the number of layers $B$,
quadratically increases with the number of examples in a task $N^{\mathrm{L}}+N^{\mathrm{U}}$,
and quadratically increases with the number of attributes and classes in a task $M+C$.
The quadratic growth is for self-attention.
When example, attribute, and/or class sizes are huge,
we can use the techniques of attention layers for 
large-scale sets~\cite{ainslie2020etc,zaheer2020big,grefenstette2019generalized}.

\begin{algorithm}[t!]
  \centering
  \caption{Meta-learning procedures of our model.}
  \label{alg:train}
  \begin{algorithmic}[1]
    \renewcommand{\algorithmicrequire}{\textbf{Input:}}
    \renewcommand{\algorithmicensure}{\textbf{Output:}}
    \REQUIRE{Meta-training data $\mathcal{D}$,
      number of labeled $N^{\mathrm{L}}$
      and unlabeled $N^{\mathrm{U}}$ examples per task.}
    \ENSURE{Trained model parameters $\bm{\Theta}$.}
    \STATE Initialize model parameters $\bm{\Theta}$.
    \WHILE{End condition is satisfied}
    \STATE Randomly select dataset index $t$ from $\{1,\cdots,T\}$.
    \STATE Randomly sample $N^{\mathrm{L}}$ examples as labeled data $\vec{X}^{\mathrm{L}}, \vec{Y}^{\mathrm{L}}$, and $N^{\mathrm{U}}$ examples as unlabeled data $\vec{X}^{\mathrm{U}}, \vec{Y}^{\mathrm{U}}$ without replacement from the $t$th dataset.
    \STATE Construct input tensor $\vec{Z}^{(1)}$ using $\vec{X}^{\mathrm{L}}, \vec{Y}^{\mathrm{L}}, \vec{X}^{\mathrm{U}}$ in Eq.~(\ref{eq:z1}).
    \STATE Update the tensor by applying MVSA layers along the first and second modes in Eqs.~(\ref{eq:z2b},\ref{eq:z2b1}).
    \STATE Obtain embeddings for examples
    $\{\vec{z}_{n}^{\mathrm{L}}\}_{n=1}^{N^{\mathrm{L}}}$,
    $\{\vec{z}_{n}^{\mathrm{U}}\}_{n=1}^{N^{\mathrm{U}}}$
    from the updated tensor.
    \STATE Calculate test loss
    $-\frac{1}{N^{\mathrm{U}}}\sum_{n=1}^{N^{\mathrm{U}}}\log p(y_{n}^{\mathrm{U}}|\vec{z}^{\mathrm{U}}_{n};\vec{X}^{\mathrm{L}},\vec{Y}^{\mathrm{L}},\vec{X}^{\mathrm{U}},\bm{\Theta})$ on held-out unlabeled data.
    \STATE Update model parameters $\bm{\Theta}$ using the gradient of the loss by a stochastic gradient method.
    \ENDWHILE
  \end{algorithmic}
\end{algorithm}

\subsection{Regression given embeddings}
\label{sec:regression}

The proposed method is also applicable when tasks are regression,
where labels are continuous values, $\vec{y}\in\mathbb{R}^{C}$.
We use regression models, such as Gaussian processes (GPs) and linear regression,
for predicting labels given embeddings instead of prototypical classifiers.
With GPs, a kernel function in the embedding space $k(\vec{z},\vec{z}')\in\mathbb{R}$ is used.
The predictive distribution for the $c$th label
of embedding $\vec{z}_{n}^{\mathrm{U}}$
is given by
\begin{align}
 p(y_{c}|\vec{z}_{n}^{\mathrm{U}};\vec{X}^{\mathrm{L}},\vec{Y}^{\mathrm{L}},\vec{X}^{\mathrm{U}},\bm{\Theta})
 =
  \mathcal{N}(\vec{k}_{n}^{\top}\vec{K}^{-1}\vec{y}_{c}^{\mathrm{L}},k(\vec{z}_{n}^{\mathrm{U}},\vec{z}_{n}^{\mathrm{U}})-\vec{k}_{n}^{\top}\vec{K}^{-1}\vec{k}_{n}),
 \label{eq:py_regression}
\end{align}
where 
$\vec{k}_{n}=[k(\vec{z}_{n}^{\mathrm{U}},\vec{z}_{1}^{\mathrm{L}}),\dots,k(\vec{z}_{n}^{\mathrm{U}},\vec{z}_{N^{\mathrm{L}}}^{\mathrm{L}})]\in\mathbb{R}^{N^{\mathrm{L}}}$
is the kernel vector between $\vec{z}_{n}^{\mathrm{U}}$ and the labeled examples,
$\vec{K}\in\mathbb{R}^{N^{\mathrm{L}}\times N^{\mathrm{L}}}$ is the kernel matrix between the labeled examples,
$\vec{y}_{c}^{\mathrm{L}}\in\mathbb{R}^{N^{\mathrm{L}}}$ is
the vector of the $c$th label for the labeled data,
and
$\mathcal{N}(\bm{\mu},\bm{\Sigma})$ is a Gaussian distribution
with mean $\bm{\mu}$ and covariance $\bm{\Sigma}$.

\section{Experiments}
\label{sec:experiments}
\subsection{Data}

We evaluated the proposed method using two datasets: Circle-Spiral and OpenML.
For each dataset, we randomly split the tasks,
where 70\% of them were used for meta-training,
10\% for meta-validation,
and the remaining for meta-test.
We averaged the accuracy on the meta-test data
over ten experiments with different splits of meta-training, validation, and test data.
For each class of a task,
we used one, three, or five labeled examples,
and 20 unlabeled examples.

\begin{figure}[t!]
  \centering
  \begin{tabular}{cc}
  \includegraphics[width=14em]{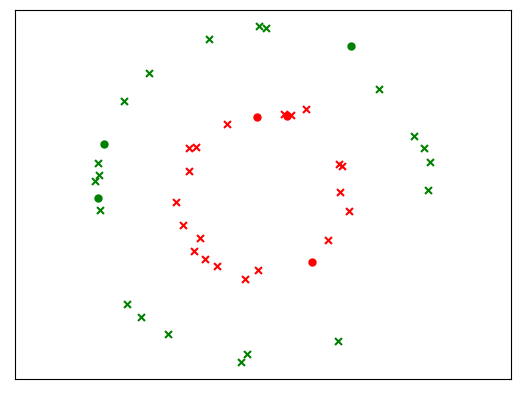}&
  \includegraphics[width=14em]{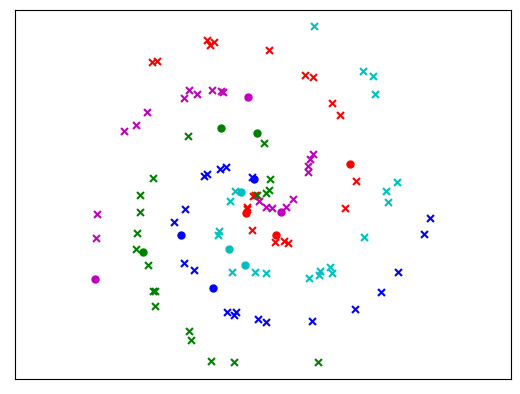}\\
  (a) Circle & (b) Spiral \\
  \end{tabular}
  \caption{Original Circle and Spiral data with three labeled and 20 unlabeled examples for each class. `o' represents a labeled example, `x' represents an unlabeled example, and the color represents the class label.} 
  \label{fig:circle-spiral}
\end{figure}

The Circle-Spiral data were synthetic with 100 tasks,
where each task was based on Circle or Spiral data.
The original Circle data consist of examples
on two concentric circles with different radius in a two-dimensional space.
The original Spiral data consist of examples
that are distributed along five spiral shaped arms.
The original Circle and Spiral data have two-dimensional attribute spaces
as shown in Figure~\ref{fig:circle-spiral}.
We transformed them into two- to ten-dimensional heterogeneous attribute spaces
by adding attributes with standard Gaussian noise
and permutating the order of the attributes for each task.
Each task contains 100 examples.

OpenML data were obtained from OpenML~\cite{vanschoren2014openml},
which is an open online platform for machine learning that holds various tasks,
using a Python API~\cite{feurer2019openml}.
We used 289 tasks in OpenML, where 
the number of attributes was between two and 1,000,
the number of classes was between two and ten,
the number of examples was between 60 and 1,000,
the number of examples per class was not less than 30,
and we omitted the tasks with the same name.
The missing values were imputed by their mean for numerical attributes
and by the most frequent value for categorical attributes.
The categorical attributes were transformed by onehot encoding.
The values were normalized in the range from zero to one.

\subsection{Compared methods}

We compared the proposed method with the following methods:
Gaussian processes (GP)~\cite{rasmussen2005gaussian},
label propagation (LP)~\cite{zhu2002learning,zhou2003learning},
model-agnostic meta-learning (MAML)~\cite{finn2017model},
prototypical networks (Proto)~\cite{snell2017prototypical},
heterogeneous meta-learning (HML)~\cite{iwata2020meta},
HML with transformer-based self-attention networks~\cite{vaswani2017attention} (AttHML),
semi-supervised learning of Proto (SemiProto) using exchangeable matrix layers (EMLs)~\cite{hartford2018deep},
meta-learning with label propagation (MetaLP)~\cite{liu2018learning},
and AttHML with label propagation (AttHMLLP).
MAML, Proto, HML, AttHML, SemiProto, MetaLP, AttHMLLP,
and the proposed method
are meta-learning schemes that use data in different tasks.
GP, and LP are not meta-learning methods
and they do not use data in different tasks
except for hyperparameter tuning.
GP, MAML, Proto, HML, and AttHML are supervised methods
that use labeled data.
LP, SemiProto, MetaLP, AttHMLLP, and the proposed method
are semi-supervised methods
that use labeled and unlabeled data.

In GP, Gaussian processes are adapted to the labeled data with onehot encoding labels for each task.
We used RBF kernels in the attribute space and the Gaussian observation noise, where
the kernel parameters were optimized using meta-training datasets.
In LP, a graph was constructed using RBF kernels in the attribute space, and
the labels were propagated through the graph~\cite{zhou2003learning} for each task.
The kernel parameters and a hyperparameter to control the amount of propagated information were optimized using meta-training datasets.
In MAML, we used neural networks based on deep sets~\cite{zaheer2017deep}
to handle data with heterogeneous attribute spaces.
The initial parameters of the neural networks were trained such that
the test loss was minimized when the parameters were adapted to the labeled data for each task.
Proto used the deep sets as in MAML to embed the data,
and the class probability was estimated with a Gaussian mixture model
in the embedding space.
MAML and Proto used
three-layered feed-forward neural networks with 32 hidden units for the deep sets.
For the inner optimization of MAML, we used five epochs of gradient descent
with learning rate $10^{-3}$.
HML is a meta-learning method for supervised learning with heterogeneous attribute space.
In HML, the embeddings of attributes, classes, and examples are obtained
using labeled data by deep sets.
The class probability is estimated with a Gaussian mixture model
in the embedding space as with Proto.
We used three-layered feed-forward neural networks with 32 hidden units,
and three iterations of deep sets.
In AttHML, transformers~\cite{vaswani2017attention} were used
for embedding instead of feed-forward neural networks in HML.
AttHML performs attentions across elements of example-attribute pairs.
AttHML used three layers of transformer encoder layers
with four heads and 32 hidden units.
SemiProto used a neural network based on exchangeable matrix layers (EMLs)~\cite{hartford2018deep}
to embed the labeled and unlabeled data with heterogeneous attributes in a task-specific space.
The neural network takes labeled and unlabeled data as input,
where the labels for the unlabeled data are treated as missing values.
Note that although EMLs have been used in meta-learning for matrix factorization~\cite{iwata2021meta},
they have not been used in meta-learning for semi-supervised learning.
MetaLP is a meta-learning method for semi-supervised learning.
We used EML-based neural networks for embedding as in SemiProto,
and estimated the class labels by label propagation using graphs based on RBF kernels in the embedding space.
In SemiProto and MetaLP, we used three layers of EMLs with 32 hidden units.
In AttHMLLP, embeddings are obtained by AttHML, and label propagation
is performed in the embedding space.

\subsection{Settings}

With the proposed method,
we used three layers of MVSAs with four heads,
where $H_{\mathrm{K}}=H_{\mathrm{V}}=H=32$ except for the last layer,
and $H^{(B)}=1$ for the last layer.
We used three-layered feed-forward neural networks with 32 hidden units
for mode-three-wise neural networks $\mathrm{FF}^{(b)}$.
We optimized our model using Adam~\cite{kingma2014adam} with learning rate $10^{-4}$,
and a batch size of eight.
The number of meta-training epochs was 5,000, and
the meta-validation data were used for early stopping.
We implemented the proposed method with PyTorch~\cite{paszke2019pytorch}.

\subsection{Results}

Table~\ref{tab:accuracy} shows the test accuracy
on the Circle-Spiral and OpenML data.
The proposed method (Ours) achieved the best performance in all cases.
As the number of labeled examples increased, the accuracy generally rose.
Since
GP, and LP are not meta-learning methods and cannot use the information on different tasks,
their accuracy was low.
MAML, Proto, HML, and AttHML
cannot use unlabeled data for obtaining task-specific classifiers.
Therefore, they underperformed the proposed method.
The better performance of the proposed method compared with the AttHML
demonstrates that our MVSA layers
are more effective than applying the existing attention layers to HML.
SemiProto, MetaLP, and AttHMLLP
can use unlabeled data as well as information on different tasks.
However, their accuracy was lower than the proposed method.
This result indicates that our MVSA layers can appropriately
learn the embeddings of the labeled and unlabeled data
with heterogeneous attribute spaces.

\begin{table}[t!]
  \centering
  \caption{Average test accuracy and its standard error on Circle-Spiral (a) and OpenML data (b) with different numbers of labeled examples per class (Shot). Values in bold are not statistically different at 5\% level from the best performing method in each case by a paired t-test.}
  \label{tab:accuracy}
  (a) Circle-Spiral data\\
  {\tabcolsep=0.3em
  \begin{tabular}{lrrr}
    \hline
    Shot & 1 & 3 & 5 \\
    \hline
GP & 0.357 $\!\pm\!$ 0.011 & 0.379 $\!\pm\!$ 0.012 & 0.401 $\!\pm\!$ 0.014 \\
LP &  0.361 $\!\pm\!$ 0.011 & 0.380 $\!\pm\!$ 0.012 & 0.396 $\!\pm\!$ 0.013 \\
MAML & 0.441 $\!\pm\!$ 0.016 & 0.486 $\!\pm\!$ 0.019 & 0.502 $\!\pm\!$ 0.019 \\
Proto & 0.461 $\!\pm\!$ 0.017 & 0.503 $\!\pm\!$ 0.019 & 0.510 $\!\pm\!$ 0.019 \\
HML & 0.536 $\!\pm\!$ 0.017 & 0.660 $\!\pm\!$ 0.018 & 0.692 $\!\pm\!$ 0.019 \\
AttHML & 0.661 $\!\pm\!$ 0.021 & 0.972 $\!\pm\!$ 0.005 & {\bf 0.991 $\!\pm\!$ 0.001} \\
SemiProto & 0.662 $\!\pm\!$ 0.021 & 0.710 $\!\pm\!$ 0.016 & 0.731 $\!\pm\!$ 0.014 \\
MetaLP & 0.344 $\!\pm\!$ 0.008 & 0.639 $\!\pm\!$ 0.037 & 0.761 $\!\pm\!$ 0.012 \\
{\small AttHMLLP} & 0.345 $\!\pm\!$ 0.009 & 0.351 $\!\pm\!$ 0.010 & 0.804 $\!\pm\!$ 0.049 \\
Ours & {\bf 0.974 $\!\pm\!$ 0.011} & {\bf 0.991 $\!\pm\!$ 0.001} & {\bf 0.992 $\!\pm\!$ 0.000}\\
 \hline
  \end{tabular}}
  \\
  \vspace{0.5em}
  (b) OpenML data\\
  {\tabcolsep=0.3em
  \begin{tabular}{lrrr}
    \hline
    Shot & 1 & 3 & 5 \\
    \hline
GP &  0.602 $\!\pm\!$ 0.008 & 0.644 $\!\pm\!$ 0.011 & 0.663 $\!\pm\!$ 0.011 \\
LP &  0.587 $\!\pm\!$ 0.010 & 0.623 $\!\pm\!$ 0.010 & 0.635 $\!\pm\!$ 0.011 \\
MAML &  0.547 $\!\pm\!$ 0.009 & 0.575 $\!\pm\!$ 0.011 & 0.583 $\!\pm\!$ 0.011 \\
Proto & 0.548 $\!\pm\!$ 0.010 & 0.575 $\!\pm\!$ 0.011 & 0.584 $\!\pm\!$ 0.011 \\
HML & 0.546 $\!\pm\!$ 0.009 & 0.582 $\!\pm\!$ 0.012 & 0.598 $\!\pm\!$ 0.012 \\
AttHML & 0.581 $\!\pm\!$ 0.011 & 0.634 $\!\pm\!$ 0.009 & 0.659 $\!\pm\!$ 0.015 \\
SemiProto & 0.607 $\!\pm\!$ 0.011 & 0.654 $\!\pm\!$ 0.010 & 0.671 $\!\pm\!$ 0.011 \\
MetaLP & 0.604 $\!\pm\!$ 0.010 & 0.644 $\!\pm\!$ 0.010 & 0.660 $\!\pm\!$ 0.011 \\
{\small AttHMLLP} & 0.527 $\!\pm\!$ 0.009 & 0.574 $\!\pm\!$ 0.010 & 0.586 $\!\pm\!$ 0.012 \\
Ours & {\bf 0.647 $\!\pm\!$ 0.010} & {\bf 0.703 $\!\pm\!$ 0.012} & {\bf 0.715 $\!\pm\!$ 0.010}\\
    \hline
 \end{tabular}}
\end{table}

\begin{figure*}[t!]
  \centering
  Circle task\\
   {\tabcolsep=0.5em\begin{tabular}{cccc}
  \includegraphics[width=10em]{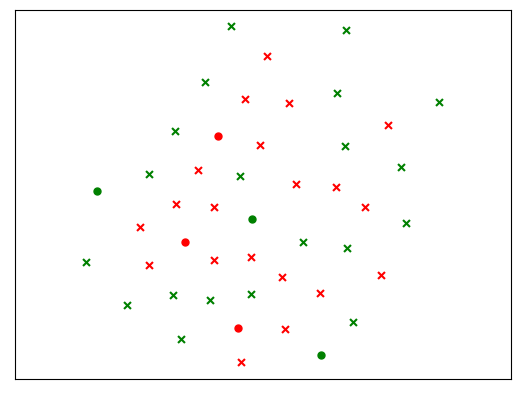}&
  \includegraphics[width=10em]{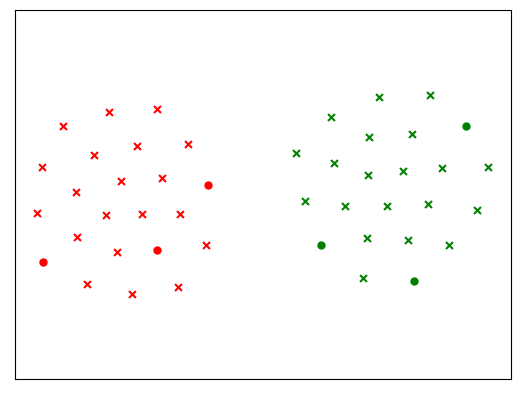}&
  \includegraphics[width=10em]{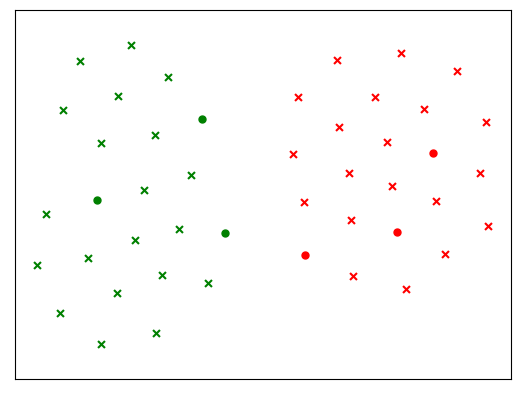}&
  \includegraphics[width=10em]{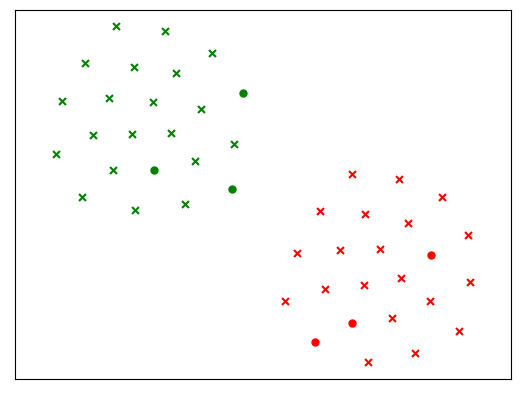}\\
   \end{tabular}}
   \\
   Spiral task\\
   {\tabcolsep=0.5em\begin{tabular}{cccc}
  \includegraphics[width=10em]{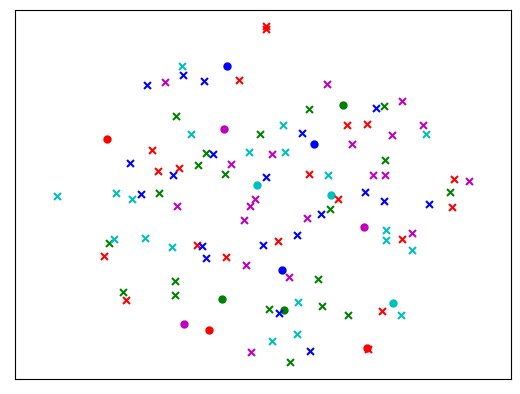}&
  \includegraphics[width=10em]{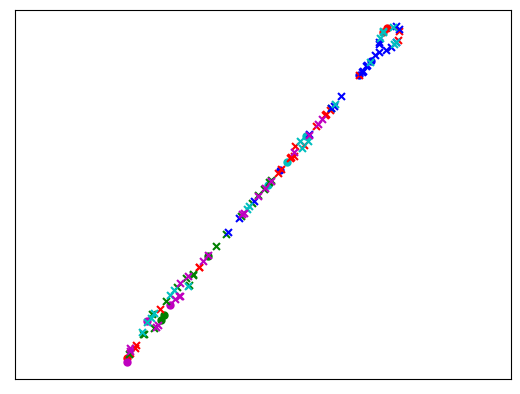}&
  \includegraphics[width=10em]{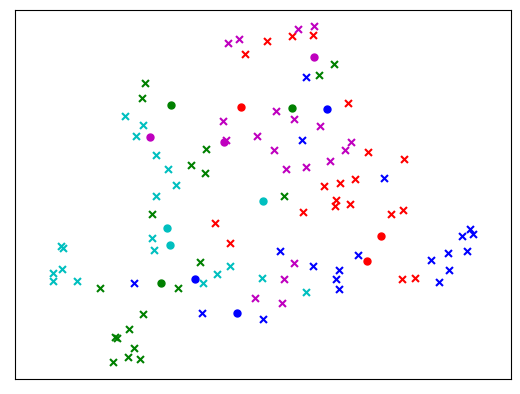}&
  \includegraphics[width=10em]{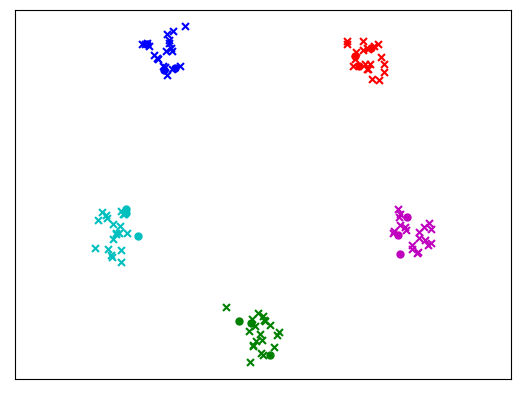}\\
  (a) Original data & (b) HML &
  (c) Proto & (d) Ours \\
  \end{tabular}}
  \caption{t-SNE~\cite{maaten2008visualizing} visualization of labeled and unlabeled examples in two tasks of Circle-Spiral data with three labeled examples per class, where the numbers of attributes of the Circle (top) and Spiral (bottom) tasks are four and six, respectively. (a) is the visualization of the original attribute vectors, and (b,c,d) are the visualization of the learned representation vectors by HML, Proto, and the proposed method.}
  \label{fig:tsne}
\end{figure*}

The efficacy of the proposed method is also
shown in the two-dimensional visualization of the original data
and the embeddings
in Figure~\ref{fig:tsne}.
Since the original data contained attributes with Gaussian noise,
there was no cluster structure of classes (a).
Although HML and Proto successfully learned the embeddings for few-shot classification on the Circle task,
they failed on the Spiral task (b,c).
On the other hand, the embeddings by the proposed method
exhibit clear class structure on both the Circle and Spiral tasks (d).

\begin{figure*}[t!]
  \centering
  {\tabcolsep=1em\begin{tabular}{ccc}
  \includegraphics[width=15em]{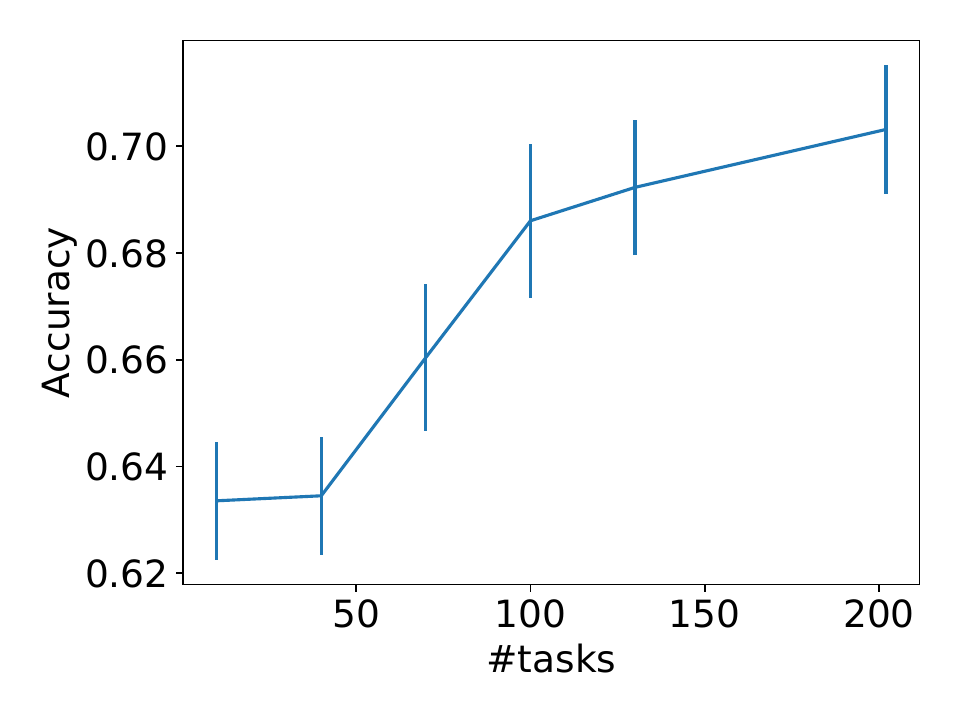}&
  \includegraphics[width=15em]{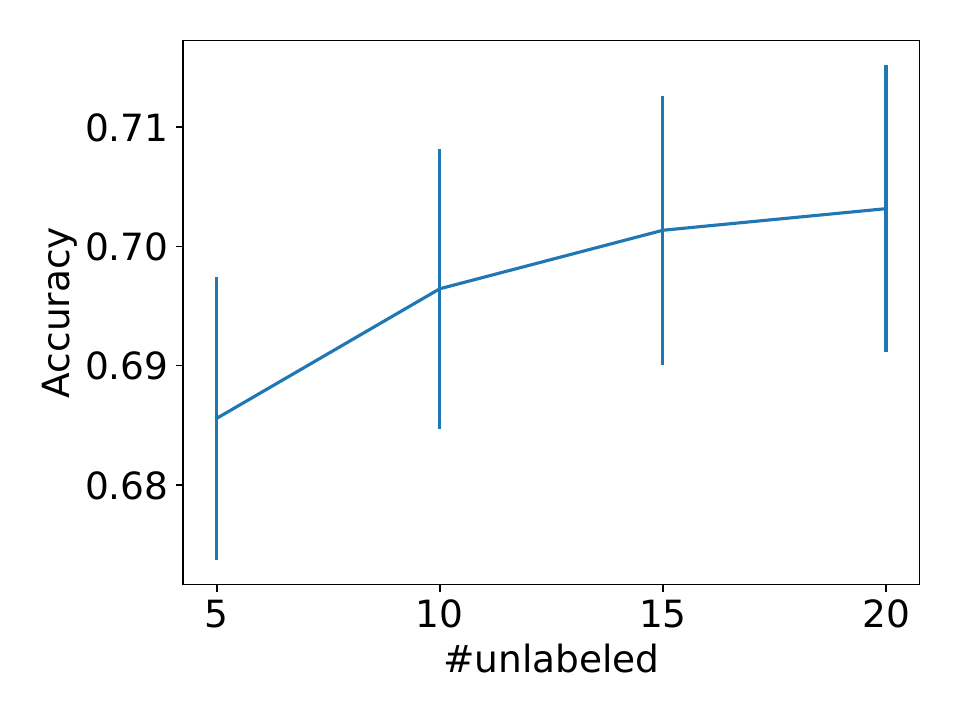}\\
  (a) \#meta-training tasks & (b) \#unlabeled examples \\
  \includegraphics[width=15em]{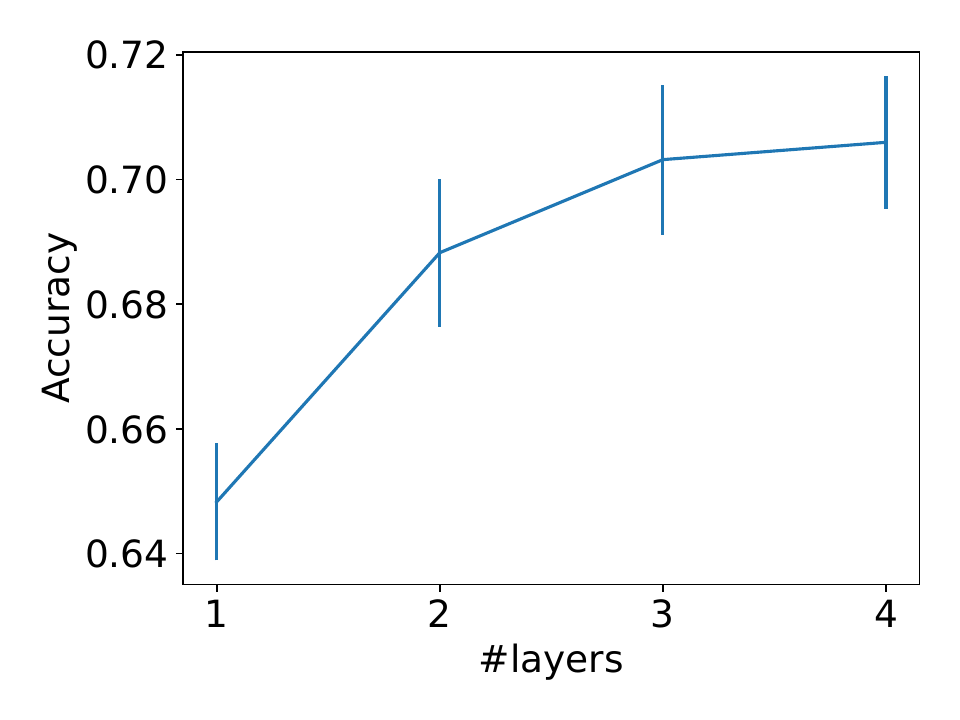}&
    \includegraphics[width=15em]{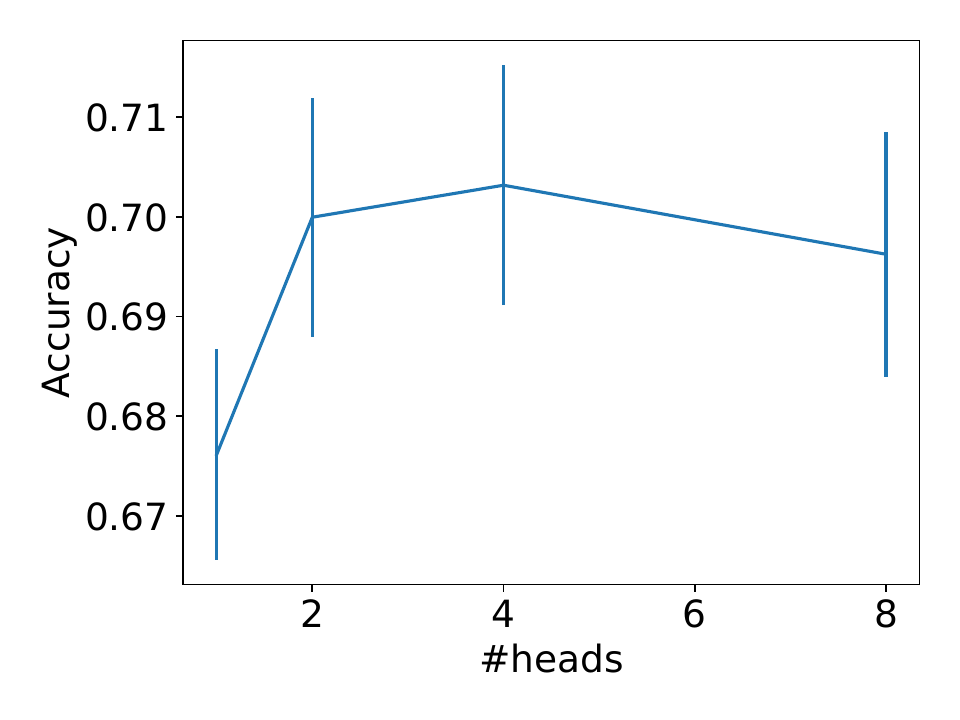}\\  
  (c) \#layers & (d) \#heads \\
  \end{tabular}}
  \caption{Average test accuracy and its standard error by the proposed method
    on OpenML data
    with different numbers of meta-training tasks (a),
    different numbers of unlabeled examples per class (b),
    different numbers of MVSA layers (c), and
    different numbers of heads (d).}
  \label{fig:acc_different}
\end{figure*}

Figure~\ref{fig:acc_different}(a) shows that the accuracy increased
as the number of meta-training tasks rose with the proposed method.
This result indicates that it is important to meta-learn from a wide variety of tasks
to improve performance.
Figure~\ref{fig:acc_different}(b) shows that the accuracy increased 
as the number of unlabeled data for each task increased with the proposed method.
This result indicates that
the proposed method improves performance by collecting unlabeled data for each task.
Figure~\ref{fig:acc_different}(c) shows that
the performance was low when the number of MVSA layers was small with the proposed method.
It is difficult to extract useful information
from the given labeled and unlabeled data with a limited number of MVSA layers.
Figure~\ref{fig:acc_different}(d) shows that
the performance was low when single-head VSA layers were used
in the proposed method.

Table~\ref{tab:ablation} shows the ablation study results.
When no attentions were performed across attributes (ExampleAttn), the accuracy was low.
This result indicates the importance of attribute-wise attentions
to obtain representations for tasks with heterogeneous attribute spaces.
When attentions across examples were omitted (AttributeAttn),
the performance was not degraded on the Circle-Spiral data.
However, it was degraded on the OpenML data
since the OpenML data contain a wide variety of tasks
and example-wise attentions help extract information about tasks.
When we did not use observation indicators in the input of our model (w/oObsInd),
the accuracy was low
because the observation indicators help distinguish between labeled and unlabeled examples,
and between attributes and labels.
On the other hand,
the elimination of attribute and label indicators (w/oAttLabInd) only slightly decreased the performance
except for the case of 1-shot on the Circle-Spiral data.
This is because observation indicators can be used to distinguish attributes and labels.
Without residual blocks (w/oRes), the training did not progress.
Without layer normalization (w/oLN), the performance deteriorated.
Residual blocks and layer normalization are commonly used for
attention models~\cite{bahdanau2015neural,vaswani2017attention,lee2019set,kossen2021self},
and they are also effective for our model.

\begin{table}[t!]
  \centering
  \caption{Ablation study.
    Average test accuracy and its standard error.
    ExampleAttn is the proposed method that performs attentions across examples with MVSA without attentions across attributes. AttributeAttn is the proposed method that performs attentions across attributes with MVSA without attentions across examples. w/oObsInd is the proposed method without observed indicators $\vec{Z}^{(1)}_{::2}$ in the input tensor. w/oAttLabInd is the proposed method without attribute and label indicators, $\vec{Z}^{(1)}_{::3}$ and $\vec{Z}^{(1)}_{::4}$, in the input tensor.
    w/oRes is the proposed method without the residual term, and
    w/oLN is the proposed method without layer normalization.}
  \label{tab:ablation}
  (a) Circle-Spiral data\\
  \begin{tabular}{lrrr}
    \hline
    Shot & 1 & 3 & 5 \\
    \hline
Ours & {\bf 0.974 $\pm$ 0.011} &{\bf 0.991 $\pm$ 0.001} &0.992 $\pm$ 0.000 \\
ExampleAttn & 0.481 $\pm$ 0.024 & 0.740 $\pm$ 0.019 & 0.852 $\pm$ 0.021 \\
AttributeAttn & {\bf 0.985 $\pm$ 0.003} & {\bf 0.988 $\pm$ 0.002} & {\bf 0.994 $\pm$ 0.000} \\
w/oObsInd & 0.780 $\pm$ 0.054 & 0.654 $\pm$ 0.065 & 0.893 $\pm$ 0.023 \\
w/oAttLabInd & 0.934 $\pm$ 0.025 & {\bf 0.990 $\pm$ 0.002} & 0.992 $\pm$ 0.001 \\
w/oRes & 0.342 $\pm$ 0.009 & 0.342 $\pm$ 0.009 & 0.342 $\pm$ 0.009 \\
w/oLN & 0.358 $\pm$ 0.010 & 0.774 $\pm$ 0.050 & 0.937 $\pm$ 0.018 \\
    \hline
  \end{tabular}
  \\
  (b) OpenML data\\
  \begin{tabular}{lrrr}
    \hline
    Shot & 1 & 3 & 5 \\
    \hline
Ours & {\bf 0.647 $\pm$ 0.010} &{\bf 0.703 $\pm$ 0.012} &{\bf 0.715 $\pm$ 0.010} \\
ExampleAttn & 0.601 $\pm$ 0.009 & 0.651 $\pm$ 0.010 & 0.680 $\pm$ 0.013 \\
AttributeAttn & 0.609 $\pm$ 0.011 & 0.677 $\pm$ 0.011 & {\bf 0.710 $\pm$ 0.011} \\
w/oObsInd & 0.617 $\pm$ 0.011 & 0.693 $\pm$ 0.013 & 0.706 $\pm$ 0.010 \\
w/oAttLabInd & {\bf 0.638 $\pm$ 0.014} & {\bf 0.695 $\pm$ 0.011} & {\bf 0.713 $\pm$ 0.011} \\
w/oRes & 0.472 $\pm$ 0.005 & 0.472 $\pm$ 0.005 & 0.472 $\pm$ 0.005 \\
w/oLN & 0.611 $\pm$ 0.009 & 0.680 $\pm$ 0.011 & 0.706 $\pm$ 0.012 \\
 \hline
  \end{tabular}
\end{table}

\begin{figure}[t!]
  \centering
  \includegraphics[width=15em]{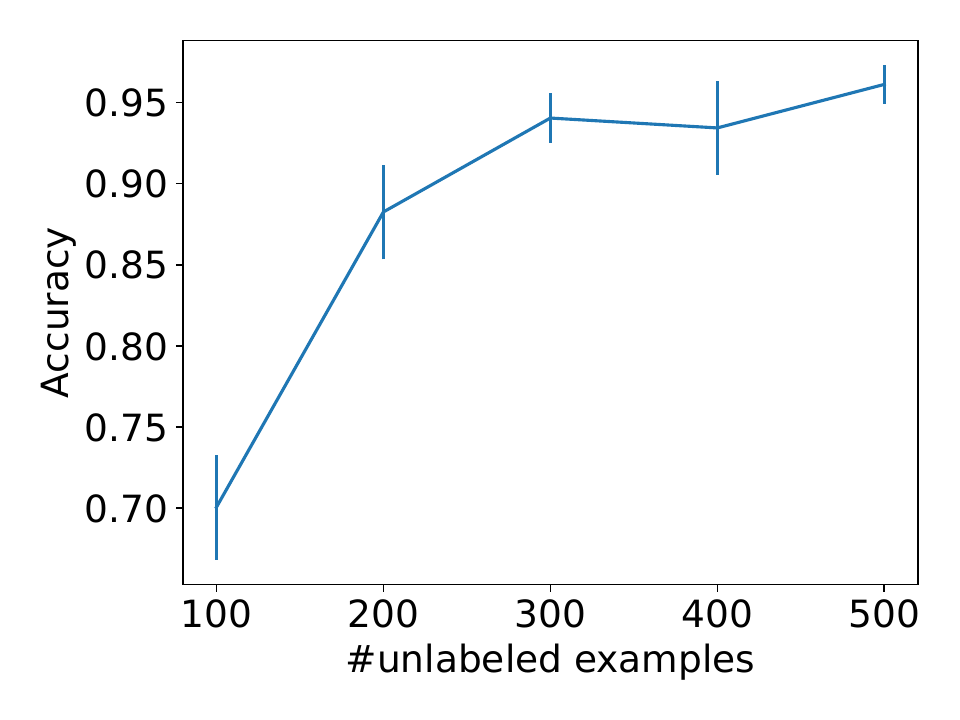}
  \caption{Average test accuracy and its standard error by the proposed method
    with different numbers of unlabeled examples in meta-training on Circle-Spiral data
    with 500 unlabeled examples in meta-test.}
  \label{fig:n_qry_train}
\end{figure}

Figure~\ref{fig:n_qry_train} shows the performance with different numbers of unlabeled examples in meta-training,
where the number of unlabeled examples in meta-test was 500,
the number of labeled examples was one, and the number of meta-training tasks was 40.
The accuracy improved as the unlabeled example size increased,
even when the unlabeled example size in meta-training was not the same as in meta-test.

Table~\ref{tab:time} shows the computational time for training
with a GTX 1080Ti GPU.
AttHML and AttHMLLP took a long time since they contained many parameters
to be learned.
MAML also took a long time since it needed
iterative gradient descent steps for each of the task-specific adaptations.
The training time with the proposed method was shorter than them,
but longer than the other methods
since it uses powerful but expensive self-attention layers for embedding.

\begin{table*}[t!]
  \centering
  \caption{Computational time in hours for training on OpenML data with three labeled examples per class.}
  \label{tab:time}
        {\tabcolsep=0.5em
          \begin{tabular}{rrrrrrrrrr}
    \hline
    GP & LP & MAML & Proto & HML & AttHML & SemiProto & MetaLP & AttHMLLP & Ours\\
    \hline
    3.0 & 2.3 & 18.6 & 2.0 & 7.6 & 50.0 & 3.1 & 4.5 & 34.4 & 9.9\\    
    \hline
  \end{tabular}}
\end{table*}

We also evaluated on the regression tasks.
We used 221 regression tasks with a single target variable in OpenML, where 
the number of attributes was between two and 1,000,
the number of examples was between 60 and 1,000,
and we omitted the tasks with the same name.
The values were normalized with zero mean and unit standard deviation.
Table~\ref{tab:mse} shows the test mean squared error.
The proposed method, MetaGP, HML,
AttHML, and SemiMetaGP used GPs with RBF kernels
in the embedding space as described in Section~\ref{sec:regression}.
For embedding neural networks,
MetaGP used deep sets, and SemiMetaGP used EMLs.
The proposed method achieved the best performance on the regression tasks.

\begin{table}[t!]
  \centering
  \caption{Average test mean squared errors and their standard errors on OpenML regression data with different numbers of labeled examples for each task (\#Labeled). Values in bold are not statistically different at 5\% level from the best performing method in each case by a paired t-test.}
  \label{tab:mse}
  {\tabcolsep=0.3em
  \begin{tabular}{lrrr}
    \hline
    \#Labeled & 10 & 15 & 20 \\
    \hline
GP & 0.899 $\!\pm\!$ 0.013 & 0.861 $\!\pm\!$ 0.016 & 0.834 $\!\pm\!$ 0.020 \\
MAML &  0.992 $\!\pm\!$ 0.013 &  0.996 $\!\pm\!$ 0.025 &  0.957 $\!\pm\!$ 0.030 \\
MetaGP &  0.989 $\!\pm\!$ 0.012 &  0.994 $\!\pm\!$ 0.025 &  0.961 $\!\pm\!$ 0.030 \\
HML &  0.763 $\!\pm\!$ 0.019 &  0.715 $\!\pm\!$ 0.025 &  0.687 $\!\pm\!$ 0.024 \\
AttHML & 0.665 $\!\pm\!$ 0.029 & {\bf  0.620 $\!\pm\!$ 0.029} &  0.581 $\!\pm\!$ 0.034 \\
{\small SemiMetaGP} & 0.845 $\!\pm\!$ 0.022 &  0.787 $\!\pm\!$ 0.027 &  0.729 $\!\pm\!$ 0.028 \\
Ours & {\bf  0.628 $\!\pm\!$ 0.030} & {\bf  0.600 $\!\pm\!$ 0.030} & {\bf  0.550 $\!\pm\!$ 0.033}\\
    \hline
  \end{tabular}}
\end{table}

\section{Conclusion}

We proposed a neural network-based meta-learning method
for semi-supervised learning that
learns from tasks with heterogeneous attribute spaces
to improve performance in unseen tasks with labeled and unlabeled data.
The proposed method achieved significantly better performance than
the existing methods.
Although we believe that our work is an important step
for learning from a wide variety of tasks,
we must extend our approach in several directions.
For future work, we plan to improve the scalability
using efficient attention layers~\cite{ainslie2020etc,zaheer2020big,grefenstette2019generalized}.
Also, we would like to apply our variable-feature attention layers to other problems than meta-learning.

\bibliographystyle{abbrv}
\bibliography{arxiv_meta_semi}

\appendix
\onecolumn

\section{Equivariance of variable-feature attention layers along the first and second modes}
\label{app:proof}

We provide a proof that the variable-feature attention layers are equivariant to a permutation
of the slices along the first and second modes.
The proof of the equivariance of the (non-variable-feature) attention layers was provided in a previous work~\cite{kossen2021self}.

\begin{definition}
  Function $f: \mathbb{R}^{D_{1} \times D_{2} \times D_{3}} \rightarrow \mathbb{R}^{D_{1} \times D_{2} \times D_{3}}$
  is mode-$n$ equivariant if for any permutation 
  $\sigma_{n}: [1,\dots,D_{n}]\rightarrow [1,\dots,D_{n}]$
  on a slice along the $n$th mode, we have for all $i$,
  $f(\vec{Z})_{i}^{n}=f(\mathrm{concat}_{n}[\vec{Z}^{n}_{\sigma^{-1}(1)},\dots,\vec{Z}^{n}_{\sigma^{-1}(D_{n})}])^{n}_{\sigma(i)}$,
  where $\mathrm{concat}_{n}$ is the concatenation in the $n$th mode, and
  $\vec{Z}^{n}_{i}$ is the $i$th slice along the $n$th mode of $\vec{Z}$.
\end{definition}

\begin{lemma}
  The mode-three product of a three-tensor is mode-one equivariant.
  \label{lemma:mode-one}
\end{lemma}

\textit{Proof.}
  Let $\sigma^{n}(\vec{Z})$ be a permutation of the slices of $\vec{Z}$ along the $n$th mode with permutation $\sigma$.
  Then we have
  \begin{align}
    (\sigma_{1}(\vec{Z})\times_{3}\vec{W})_{ijk}
    = \sum_{d_{3}=1}^{D_{3}}z_{\sigma^{-1}(i)jd_{3}}w_{kd_{3}}
    =(\vec{Z}\times_{3}\vec{W})_{\sigma^{-1}(i)jk}    
    =\sigma_{1}(\vec{Z}\times_{3}\vec{W})_{ijk}.
  \end{align}
\hfill$\qed$

\begin{lemma}
  The mode-three product of a three-tensor is mode-two equivariant.
  \label{lemma:mode-two}  
\end{lemma}
\textit{Proof.}
  In a similar way to the proof of the mode-one equivariance,
  \begin{align}
    (\sigma_{2}(\vec{Z})\times_{3}\vec{W})_{ijk}
    = \sum_{d_{3}=1}^{D_{3}}z_{i\sigma^{-1}(j)d_{3}}w_{kd_{3}}
    =(\vec{Z}\times_{3}\vec{W})_{i\sigma^{-1}(j)k}    
    =\sigma_{2}(\vec{Z}\times_{3}\vec{W})_{ijk}.
  \end{align}
  \hfill$\qed$
  
\begin{theorem}
  Variable-feature self-attention $\mathrm{VSA}(\vec{Z})$ is mode-one equivariant.
  \label{theorem:vsa1}
\end{theorem}
\textit{Proof.}
  The product between the permutated query and key tensors is given by
  \begin{align}
    ((\sigma_{1}(\vec{Z})\times_{3} \vec{W}^{\mathrm{Q}})_{(1)}
    (\sigma_{1}(\vec{Z})\times_{3} \vec{W}^{\mathrm{K}})_{(1)}^{\top})_{ij}
    =
    (\sigma_{1}(\vec{Q})_{(1)}
    \sigma_{1}(\vec{K})_{(1)}^{\top})_{ij}
    \nonumber\\
    =
    \sum_{d_{2}=1}^{D_{2}}\sum_{h=1}^{H_{\mathrm{K}}}q_{\sigma^{-1}(i)d_{2}h}k_{\sigma^{-1}(j)d_{2}h}
    =
    (\vec{Q}_{(1)}\vec{K}_{(1)}^{\top})_{\sigma^{-1}(i)\sigma^{-1}(j)},
    \label{eq:qk}
  \end{align}
  where we used Lemma~\ref{lemma:mode-one} in the first equality.
  The softmax operation is permutation-equivariant,
  \begin{align}
    \mathrm{softmax}(\sigma_{1}(\vec{Q}_{(1)}\vec{K}_{(1)}^{\top})/\sqrt{D_{2}H_{\mathrm{K}}})_{ij}
    =
    \sigma_{1}(\mathrm{softmax}(\vec{Q}_{(1)}\vec{K}_{(1)}^{\top}/\sqrt{D_{2}H_{\mathrm{K}}}))_{ij}.
    \label{eq:softmax}    
  \end{align}
  Let $\vec{A}=\mathrm{softmax}(\vec{Q}_{(1)}\vec{K}_{(1)}^{\top}/\sqrt{D_{2}H_{\mathrm{K}}})$.
  The VSA on permutated tensor $\sigma_{1}(\vec{Z})$ becomes
  \begin{align}
    \mathrm{VSA}(\sigma_{1}(\vec{Z}))_{ijk}
    =
    (\sigma_{1}(\vec{Z})\vec{W}^{\mathrm{V}}
    \times_{1} \sigma_{1}(\vec{A}))_{ijk}
    =
    (\sigma_{1}(\vec{V})
    \times_{1} \sigma_{1}(\vec{A}))_{ijk}
    \nonumber\\
    =
    \sum_{d_{1}=1}^{D_{1}}a_{\sigma^{-1}(i)\sigma^{-1}(d_{1})}v_{\sigma^{-1}(d_{1})jk}
    =
    (\vec{V}\times_{1}(\vec{A}))_{\sigma^{-1}(i)jk}
    =
    \sigma_{1}(\vec{V}\times_{1}(\vec{A}))_{ijk},
  \end{align}
  where we used Eqs.~(\ref{eq:qk},\ref{eq:softmax}) in the first equality,
  and Lemma~\ref{lemma:mode-one} in the second equality.
\hfill$\qed$

\begin{theorem}
  Variable-feature self-attention $\mathrm{VSA}(\vec{Z})$ is mode-two equivariant.
  \label{theorem:vsa2}
\end{theorem}

\textit{Proof.}
  The proof resembles that of Theorem~\ref{theorem:vsa1}
  except that we use Lemma~\ref{lemma:mode-two} instead of Lemma~\ref{lemma:mode-one}.
\hfill$\qed$

\begin{theorem}
  Multi-head variable-feature self-attention $\mathrm{MVSA}(\vec{Z})$ is mode-one equivariant.
  \label{theorem:mvsa1}
\end{theorem}

\textit{Proof.}
  The MVSA on permutated tensor $\sigma_{1}(\vec{Z})$ becomes
  \begin{align}
    \mathrm{MVSA}(\sigma_{1}(\vec{Z}))_{ijk}&=(\mathrm{concat}(\mathrm{VSA}_{1}(\sigma_{1}(\vec{Z})),\dots,\mathrm{VSA}_{R}(\sigma_{1}(\vec{Z})))
    \times_{3}\vec{W}^{\mathrm{O}})_{ijk}
    \nonumber\\
  &=
  (\mathrm{concat}(\sigma_{1}(\mathrm{VSA}_{1}(\vec{Z})),\dots,\sigma_{1}(\mathrm{VSA}_{R}(\vec{Z}))
    \times_{3}\vec{W}^{\mathrm{O}})_{ijk}
    \nonumber\\
    &=
    \sum_{d_{3}=1}^{D_{3}}\sum_{r=1}^{R}o_{r\sigma^{-1}(i)jd_{3}}w^{\mathrm{O}}_{rkd_{3}}
    \nonumber\\
    &=
    (\mathrm{concat}(\mathrm{VSA}_{1}(\vec{Z}),\dots,\mathrm{VSA}_{R}(\vec{Z}))
  \times_{3}\vec{W}^{\mathrm{O}})_{\sigma^{-1}(i)jk}  
    \nonumber\\
    &=
    \sigma_{1}(\mathrm{concat}(\mathrm{VSA}_{1}(\vec{Z}),\dots,\mathrm{VSA}_{R}(\vec{Z}))
    \times_{3}\vec{W}^{\mathrm{O}})_{ijk},
  \end{align}
  where we used Theorem~\ref{theorem:vsa1} in the second equality,
  $o_{rijk}=\mathrm{VSA}_{r}(\vec{Z})_{ijk}$,
  and $w_{rkk'}=\vec{W}^{\mathrm{O}}_{rH_{\mathrm{K}}+k,k'}$.
\hfill$\qed$

\begin{theorem}
  Multi-head variable-feature self-attention $\mathrm{MVSA}(\vec{Z})$ is mode-two equivariant.
  \label{theorem:mvsa2}
\end{theorem}
\textit{Proof.}
  The proof resembles that of Lemma~\ref{theorem:mvsa1}.
\hfill$\qed$

\section{Embedding models}
\label{app:embedding}

\begin{figure}[t!]
  \centering
  {\tabcolsep=3em\begin{tabular}{cc}
  \includegraphics[width=14em]{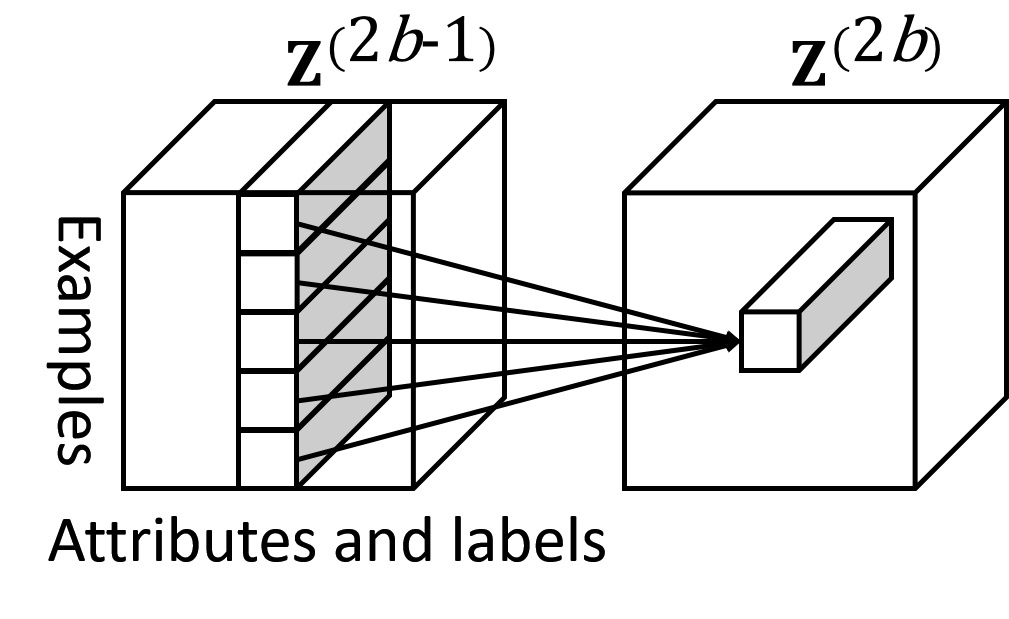}&
  \includegraphics[width=14em]{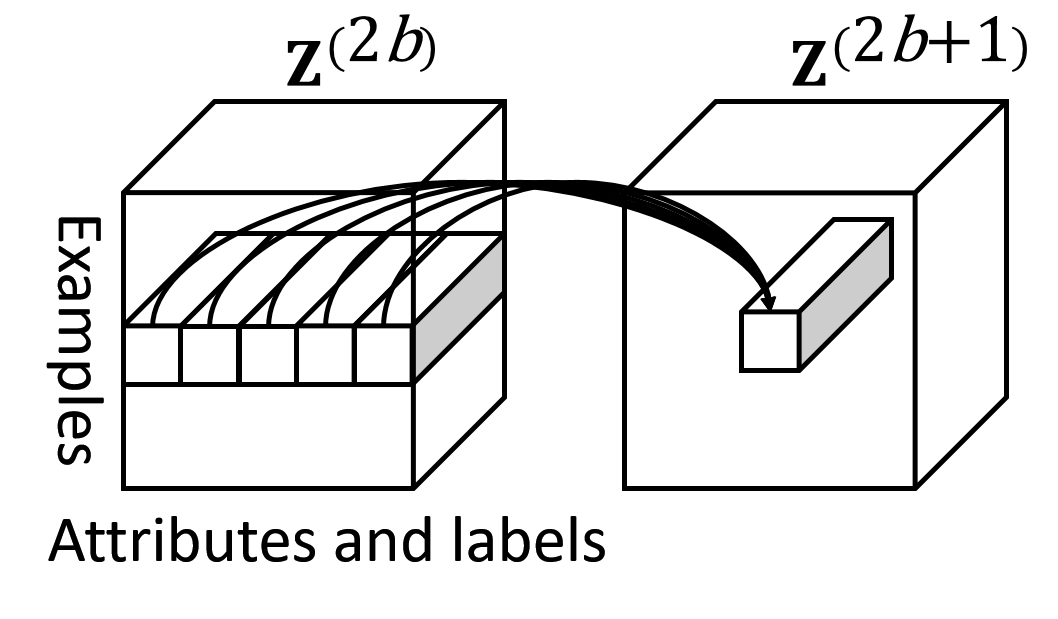}\\
  (a) Even number block & (b) odd number block\\
  \end{tabular}}
  \caption{(a) Embeddings at the even number block are obtained depending on previous embeddings of the same attribute or label of all examples with our model. (b) Embeddings at the odd number blocks are obtained depending on previous embeddings of all attributes and labels of the example.}
  \label{fig:attention}
\end{figure}

The embeddings of the $m$th attribute of the $n$th example
at the even number blocks
is given by
\begin{align}
\vec{z}_{nm}^{(2b)}=\vec{W}^{\mathrm{R}(2b-1)\top}\vec{z}_{nm}^{(2b-1)}
+\mathrm{FF}^{(2b-1)}\left(\mathrm{LN}^{(2b-1)}\left(\sum_{n'=1}^{N^{\mathrm{L}}+N^{\mathrm{U}}}a_{nn'}\vec{W}^{\mathrm{V}(2b-1)\top}\vec{z}_{n'm}^{(2b-1)}
\right)\right),
\end{align}
using Eqs.~(\ref{eq:qkv},\ref{eq:f},\ref{eq:z2b}),
where $a_{nn'}$ is the attention weight between the $n$th and $n'$th examples.
Here, we assume single-head VSA layers for simplicity.
It shows that the embedding
is calculated depending on
the embeddings of all examples $\{\vec{z}_{n'm}^{(2b-1)}\}_{n'=1}^{N^{\mathrm{L}}+N^{\mathrm{U}}}$
at the previous block
as shown in Figure~\ref{fig:attention}(a).
The embeddings of the $m$th attribute of the $n$th example
at the odd number blocks,
is given by
\begin{align}
\vec{z}_{nm}^{(2b+1)}=\vec{W}^{\mathrm{R}(2b)\top}\vec{z}_{nm}^{(2b)}
+\mathrm{FF}^{(2b)}\left(\mathrm{LN}^{(2b)}\left(\sum_{m'=1}^{M+C}a_{mm'}\vec{W}^{\mathrm{V}(2b)\top}\vec{z}_{nm'}^{(2b)}
\right)\right),
\end{align}
using Eqs.~(\ref{eq:qkv},\ref{eq:f},\ref{eq:z2b1}),
where $a_{mm'}$ is the attention weight between the $m$th and $m'$th attributes or classes.
It shows that the embedding
is calculated depending on
the embeddings of all attributes and classes $\{\vec{z}_{nm'}^{(2b)}\}_{m'=1}^{M+C}$
at the previous block
as shown in Figure~\ref{fig:attention}(b).
Although the VSA layer applies the same projection weights
for all examples, attributes, classes, and tasks,
since the values are different across examples, attributes, classes, and tasks,
the VSA layer can output example-, attribute-, class-, and task-specific embeddings
depending on the given labeled and unlabeled data.

\end{document}